\documentclass[11pt,a4paper]{article}
\usepackage[T1]{fontenc}

\usepackage[a4paper,top=2cm,bottom=2cm,left=3cm,right=3cm,marginparwidth=1.75cm]{geometry}
 
\usepackage{amsmath}
\usepackage{graphicx}
\usepackage[hidelinks]{hyperref}

\usepackage[affil-it]{authblk}

\makeatletter
\renewcommand{\section}{\@startsection {section}{1}{\z@}%
              {24pt}{12pt} {\large\scshape\bfseries}}

\renewcommand{\subsection}{\@startsection {subsection}{2}{\z@}%
             {12pt}{12pt}  {\itshape\bfseries}}

\usepackage{apacite}
\usepackage{natbib}

\title{\bfseries \normalsize Guided Persona-based AI Surveys: \\Can we replicate personal mobility preferences at scale using LLMs?}

\author{Ioannis Tzachristas, Santhanakrishnan Narayanan and Constantinos Antoniou}

\affil{Chair of Transportation Systems Engineering, TUM School of Engineering and Design, TUM, Germany}

\date{\vspace{-5ex}}

\begin{document}
\maketitle

\section*{Short summary}\small
This study explores the potential of Large Language Models (LLMs) to generate artificial surveys, with a focus on personal mobility preferences in Germany. By leveraging LLMs for synthetic data creation, we aim to address the limitations of traditional survey methods, such as high costs, inefficiency and scalability challenges. A novel approach incorporating "Personas" - combinations of demographic and behavioural attributes - is introduced and compared to five other synthetic survey methods, which vary in their use of real-world data and methodological complexity. The MiD 2017 dataset, a comprehensive mobility survey in Germany, serves as a benchmark to assess the alignment of synthetic data with real-world patterns. The results demonstrate that LLMs can effectively capture complex dependencies between demographic attributes and preferences while offering flexibility to explore hypothetical scenarios. This approach presents valuable opportunities for transportation planning and social science research, enabling scalable, cost-efficient and privacy-preserving data generation.

\textbf{Keywords}: artificial surveys, LLMs, mobility preferences, synthetic data, Persona-based

\section{Introduction}
Traditional survey methods face significant challenges, including high costs, inefficiencies and scalability limitations. As transportation systems evolve, there is an increasing demand for more efficient and adaptable data collection methods. In response, this study investigates the feasibility of using Large Language Models (LLMs) to enhance travel survey data collection \citep{liu2024bestpracticeslessonslearned, tan2024largelanguagemodelsdata}.

\begin{figure}[h!]
    \centering
    \includegraphics[width=0.97\linewidth]{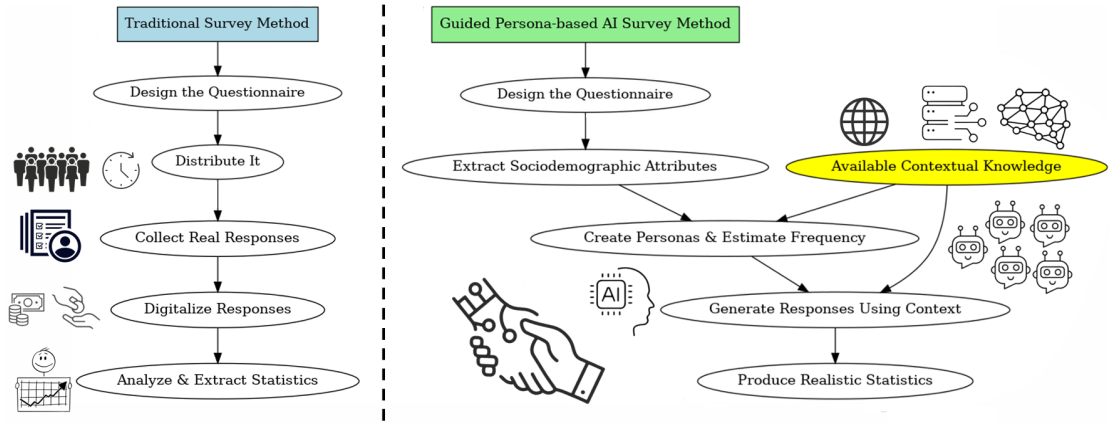}
    \caption{Traditional Survey Method vs Guided Persona-based AI Survey Method}
    \label{fig:enter-label}
\end{figure}

We propose a novel guided Persona-based AI survey methodology\footnote{\url{https://github.com/jtzach/LLM-based-AI-surveys-MiD2017.git}}, which, to the best of our knowledge, has not been previously applied to travel surveys. This approach systematically integrates sociodemographic and behavioural attributes to generate synthetic survey responses. To evaluate its effectiveness, we benchmark our method against other synthetic survey generation approaches using the MiD 2017 dataset \citep{mid2017}, a comprehensive national mobility survey in Germany. Our findings demonstrate that the guided Persona-based approach significantly outperforms competing methods in terms of accuracy and consistency with observed mobility behaviours and patterns.

Existing studies in synthetic data generation have primarily focused on demographic simulation or imputation techniques, with limited applications to large-scale survey replication \citep{wang2024surveydatasynthesisaugmentation, ma2024algorithmicfidelitylargelanguage, 10366424, Bisbee_Clinton_Dorff_Kenkel_Larson_2024}. While these methods effectively replicate population structures, they often fail to capture intricate dependencies between demographic attributes and travel behaviours. For example, economic status may influence preferences for cycling or public transport, but existing methods may overlook such correlations unless explicitly predefined. Our guided Persona-based approach addresses these limitations by leveraging LLMs' contextual understanding to generate more context-aware and realistic synthetic data.

This study contributes to the advancement of artificial survey methodologies by offering a scalable, cost-effective and privacy-preserving alternative to traditional survey methods. The results highlight the potential of guided Persona-based AI surveys to support transportation planning and social science research by enabling the creation of high-fidelity synthetic datasets.

\subsection{Why LLM-generated Artificial Surveys?}
Artificial surveys offer numerous advantages over traditional survey methods:
\begin{itemize}
    \item \textbf{Privacy-preservation}: By generating synthetic data, researchers can analyze population-level trends without accessing sensitive individual data \citep{livieris2024evaluationframeworksyntheticdata}.
    \item \textbf{Cost-efficiency}: Synthetic data generation reduces the need for repeated data collection, significantly lowering costs.
    \item \textbf{Scalability}: LLMs can generate large datasets that encompass diverse scenarios, enabling analyses that were previously unfeasible \citep{tan2024largelanguagemodelsdata}.
    \item \textbf{Flexibility}: Researchers can simulate responses to questions that were not included in the original survey.
    \item \textbf{Adaptability}: Artificial surveys can be tailored to model specific sub-populations or explore "what-if" scenarios.
\end{itemize}

These benefits demonstrate the transformative potential of synthetic data in advancing survey methodologies and addressing long-standing challenges in social science research.

\section{Methodology}

\subsection{Dataset Overview}
The MiD 2017 dataset captures detailed mobility preferences in Germany, including demographic distributions, transport modes and trip frequencies. Key attributes include age groups, education levels, main activities, economic status and household types. We normalize the translated to English version of the dataset so that the percentage of "not specified" option is merged to the most popular responses of each question as shown in the Appendix (\textbf{Figure \ref{fig:MiD 2017 dataset}}). 

\subsection{Survey Generation Methods}

For AI-survey generation, the GPT-4o API was utilized in a framework inspired by previous work \citep{tzachristas2024creatingllmbasedaiagenthighlevel}. Synthetic populations of 10,000 individuals were generated for all non-Persona-based methods. Persona-based methods involved creating distinct demographic combinations, leading to a total of 15,840 unique Personas, each defined by combinations of attributes such as age group, education level, main activity, economic status and household type.

\begin{figure}[h!]
    \centering
    \includegraphics[width=1.0\linewidth]{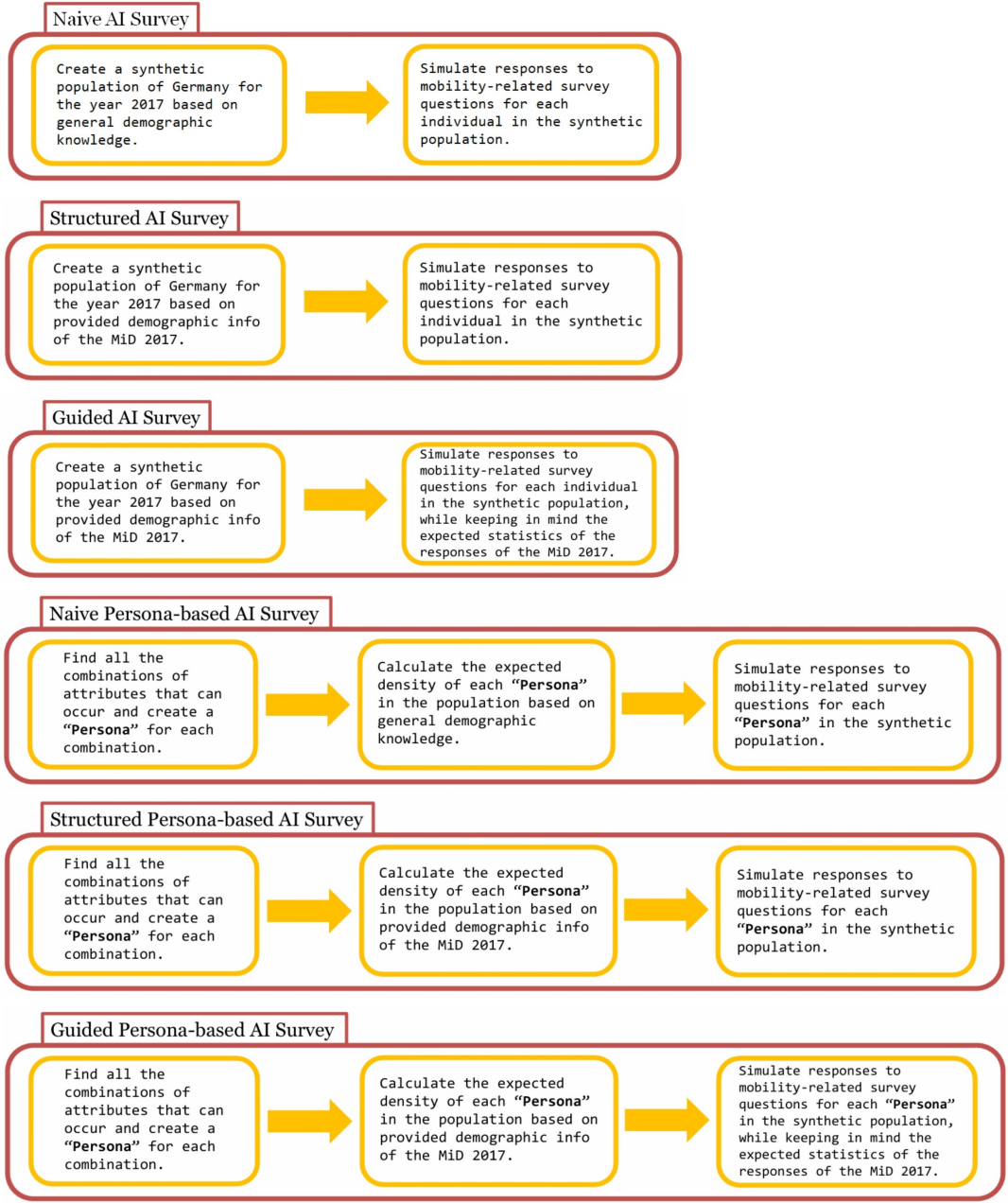}
    \caption{Overview of the Survey Generation Methods}
    \label{fig:AI-Survey methods}
\end{figure}

The survey generation techniques employed a range of prompts tailored to each method. The prompts used for each method, as well as the LLM-generated code of the experiments are available here: \url{https://github.com/jtzach/LLM-based-AI-surveys-MiD2017.git}. \\\textbf{Figure \ref{fig:AI-Survey methods}} provides an overview of the methods. 
\\\\

\textbf{Naive AI-survey}: \\A synthetic population for 2017 Germany was generated based on general demographic knowledge. Mobility-related survey responses were then simulated for each individual without any constraints or alignment with real-world benchmarks \citep{Bisbee_Clinton_Dorff_Kenkel_Larson_2024}.

\textbf{Structured AI-survey}: \\The synthetic population was aligned with demographic benchmarks provided by the MiD 2017 dataset. This method ensured a realistic representation of population structures, but responses were not explicitly aligned to the observed distributions of the MiD 2017.

\textbf{Guided AI-survey}: \\This approach added constraints to both population structure and response averages, by considering the expected statistics of MiD 2017 responses.

\textbf{Naive Persona-based AI Survey}: \\All possible combinations of demographic and behavioural attributes were identified to define Personas. Densities for each Persona were estimated using general demographic knowledge. Mobility-related responses were then simulated for each Persona.

\textbf{Structured Persona-based AI Survey}: \\This method enhanced the Naive Persona-based AI Survey by aligning Persona densities with MiD 2017 demographic data.

\textbf{Guided Persona-based AI Survey}: \\Building upon the Structured Persona-based AI Survey, this method incorporated expected response statistics from the MiD 2017 dataset.

The prompts used in the experiments were specifically crafted to guide GPT-4o's generation of realistic populations and survey responses. These prompts ranged from straightforward instructions (e.g. "Generate a population based on general demographic knowledge") to advanced directives that explicitly instructed the LLM to consider correlations between attributes (e.g. "Simulate mobility preferences while maintaining realistic correlations between age, household type and economic status based on MiD 2017 data"). This approach ensured high fidelity in response generation and allowed each method to address specific research goals while capturing intricate interdependencies between attributes.

\subsection{Persona Methods for Artificial Survey Generation}

In this section, we describe the Persona-based methods used in generating artificial surveys. This method revolves around defining a set of Personas based on a combination of demographic and behavioural attributes. We construct a partition of the population into Personas. We then calculate the population percentages of these Personas and use them for weighted analyses. Below, we formally define the key concepts and methods.

\subsubsection{Definition of a Persona}
A Persona is a tuple defined by a specific combination of demographic and behavioural attributes. Let the attributes be:
\begin{itemize}
    \item \textbf{Age Group} ($A$): \{14--17, 18--29, 30--39, 40--49, 50--59, 60--64, 65--74, 75--79, 80+\}
    \item \textbf{Education Level} ($E$): \{No Degree (yet), Low, Medium, High\}
    \item \textbf{Main Activity} ($M$): \{Full-time employee, Part-time employee, Employed (unspecified), Pupil, Student, Housewife/Househusband, Pensioner, Other\}
    \item \textbf{Economic Status} ($S$): \{Very Low, Low, Medium, High, Very High\}
    \item \textbf{Household Type} ($H$): \{Young singles, Middle-aged singles, Older singles, Young two-person households, Middle-aged two-person households, Older two-person households, Households with at least 3 adults, Households with at least 1 child under 6, Households with at least 1 child under 14, Households with at least 1 child under 18, Single parents\}
\end{itemize}

A Persona $P$ is then formally defined as:
\begin{equation}
P(A, E, M, S, H),
\end{equation}
where $A \in \text{Age Group}$, $E \in \text{Education Level}$, $M \in \text{Main Activity}$, $S \in \text{Economic Status}$ and $H \in \text{Household Type}$.

\subsubsection{Population Percentage of a Persona}

The percentage of a Persona in the population is defined by accounting for the interdependencies among attributes using conditional probabilities. Let:

\begin{itemize} \item $P_A(a)$ denote the probability of an individual belonging to age group $a \in A$. \item $P_E(e \mid a)$ denote the probability of an individual having education level $e \in E$ given their age group $a$. \item $P_M(m \mid a, e)$ denote the probability of an individual engaging in main activity $m \in M$ given their age group $a$ and education level $e$. \item $P_S(s \mid a, e, m)$ denote the probability of an individual having economic status $s \in S$ given their age group $a$, education level $e$ and main activity $m$. \item $P_H(h \mid a, e, m, s)$ denote the probability of an individual living in household type $h \in H$ given their age group $a$, education level $e$, main activity $m$ and economic status $s$. \end{itemize}

The percentage of Persona $P(a, e, m, s, h)$ in the population, denoted as $\pi_P$, is calculated as: \begin{equation} \pi_P = P_A(a) \cdot P_E(e \mid a) \cdot P_M(m \mid a, e) \cdot P_S(s \mid a, e, m) \cdot P_H(h \mid a, e, m, s), \end{equation} where $a \in A$, $e \in E$, $m \in M$, $s \in S$ and $h \in H$.

This approach ensures that the correlations among attributes are respected, providing a more accurate representation of the population. The sum of $\pi_P$ over all possible Personas equals 1, maintaining a complete representation of the population: \begin{equation} \sum_P \pi_P = 1. \end{equation}

\subsubsection{Weighted Sums for Walking Preferences by Age}
The walking preferences by age are calculated as a weighted sum of responses across Personas, where the weights are the population percentages of the Personas. Let:
\begin{itemize}
    \item $R_i(a)$ be the response for walking preference $i$ (e.g. "Completely Agree") for age group $a \in A$.
    \item $\pi_P$ be the population percentage of Persona $P$.
    \item $R_i(P)$ be the probability that Persona $P$ selects response $i$.
\end{itemize}

The weighted response for walking preference $i$ for age group $a$ is:
\begin{equation}
R_i(a) = \sum_{P \in A=a} \pi_P \cdot R_i(P),
\end{equation}
where $P \in A=a$ indicates summation over all Personas whose age group is $a$. The responses for each age group are normalized to ensure they sum to 1:
\begin{equation}
\sum_i R_i(a) = 1, \quad \forall a \in A.
\end{equation}

This method ensures that the preferences reflect both the demographic structure and the behavioural tendencies of the population.

\section{Results and discussion}
\subsection{Evaluation Metrics}
We evaluate the surveys based on the calculated percentages of each response for each attribute. The surveys were evaluated using:
\begin{itemize}
    \item Statistical alignment with MiD benchmarks (distribution comparison).
    \item Error metrics: Mean Absolute Error (MAE) and Root Mean Square Error (RMSE).
    \item Jensen-Shannon (JS) Distance: A symmetric and normalized measure of divergence between distributions.
    \item Entropy: Measures the variability or uncertainty in a set of outcomes.
    \item Conditional Entropy: Assesses the remaining uncertainty in survey responses given demographic information.
    \item Cramér’s V: Measures the strength of association between variables in real and synthetic data.
\end{itemize}

Table \ref{tab:advanced_metrics} summarizes the advanced statistical metrics for all survey methods.

\begin{table}[h!]
\centering
\caption{Advanced Statistical Metrics for AI-Generated Surveys}
\label{tab:advanced_metrics}
\resizebox{\textwidth}{!}{%
\begin{tabular}{lcccccc}
\hline
\textbf{Survey Type} & \textbf{MAE} & \textbf{RMSE} & \textbf{JS Distance} & \textbf{Entropy} & \textbf{|Conditional Entropy|} & \textbf{Cramér's V} \\
\hline
Naive AI Survey & 10.53 & 11.35 & 0.19 & 3.54 & 0.24 & 0.35 \\
Structured AI Survey & 6.69 & 7.39 & 0.13 & 3.48 & 0.18 & 0.74 \\
Guided AI Survey & 3.37 & 4.65 & 0.067 & 3.28 & 0.014 & 0.68 \\
Naive Persona-based AI Survey & 9.02 & 10.28 & 0.157 & 3.48 & 0.19 & 0.73 \\
Structured Persona-based AI Survey & 8.53 & 9.73 & 0.152 & 3.48 & 0.18 & 0.73 \\
Guided Persona-based AI Survey & 0.03 & 0.17 & 0.0016 & 3.30 & 0.0005 & 1.00 \\
\hline
\end{tabular}%
}
\end{table}

\begin{figure}[h!]
\centering
\includegraphics[width=0.9\textwidth]{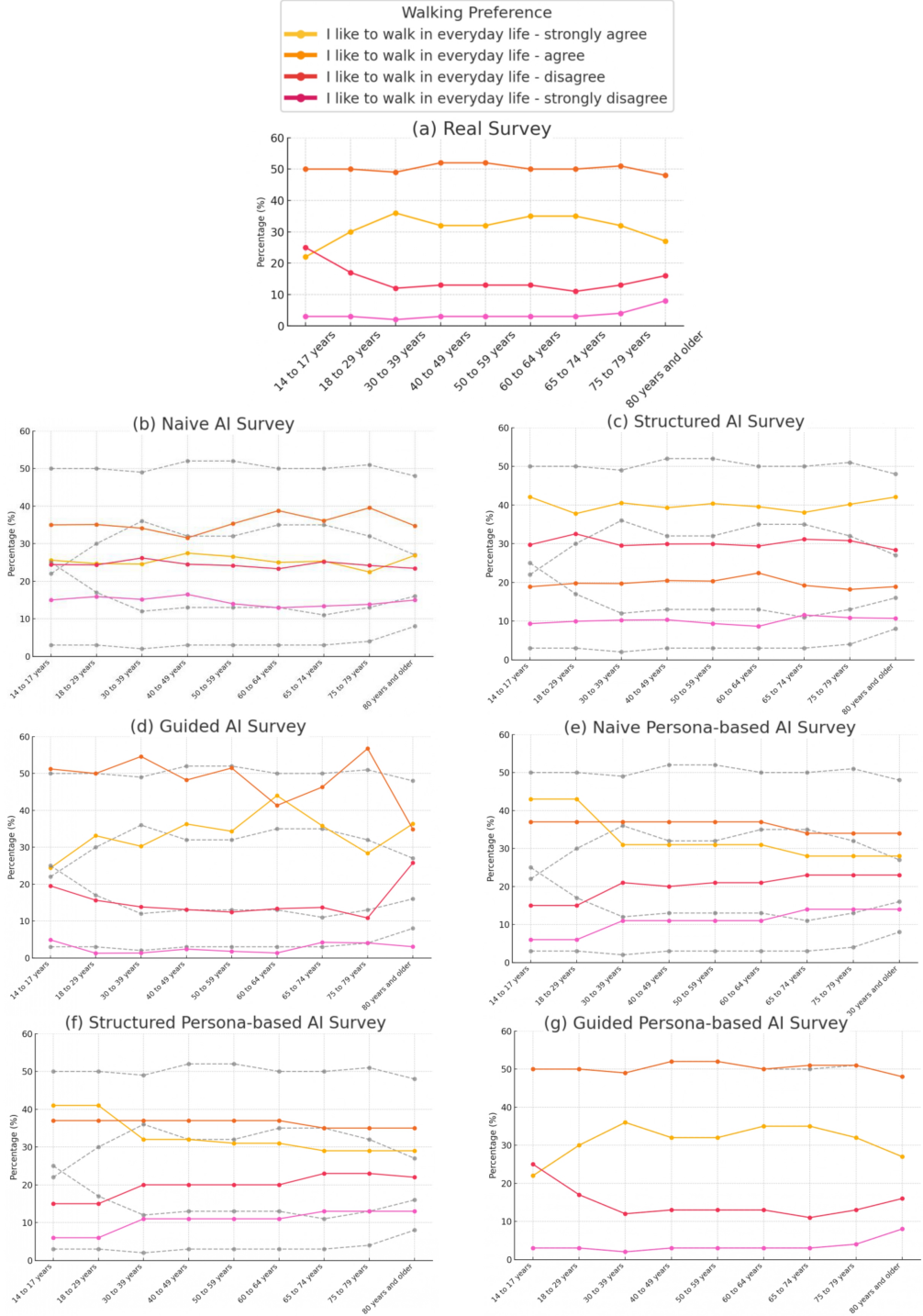}
\caption{Comparative visualization of walking preferences by age group across real and synthetic surveys. Subplots: (a) Real Survey, (b) Naive AI Survey, (c) Structured AI Survey, (d) Guided AI Survey, (e) Naive Persona-based AI Survey, (f) Structured Persona-based AI Survey, (g) Guided Persona-based AI Survey.}
\label{fig:survey_plots}
\end{figure}

\subsection{Key Findings}
\begin{itemize}
    \item \textbf{Guided Persona-based AI Survey} exhibits the most outstanding performance among all methods. With the exceptionally low MAE (0.03), RMSE (0.17) and JS Distance (0.0016), it achieves near-perfect alignment with real survey data. These metrics highlight its unparalleled ability to model demographic-response dependencies effectively.
    \item \textbf{Guided AI Survey} also performs robustly, demonstrating relatively low error metrics (MAE = 3.37, RMSE = 4.65, JS Distance = 0.067). Although slightly less precise than the Guided Persona-based AI Survey, it offers a scalable solution with strong alignment across demographic patterns.
    \item \textbf{Structured Persona-based AI Survey} displays moderate performance with higher MAE (8.53) and RMSE (9.73) values. However, its Cramér’s V value of 0.73 indicates a reasonable capacity to capture demographic patterns, albeit with greater variability.
    \item \textbf{Naive AI Survey} and \textbf{Structured AI Survey} exhibit limited alignment with real survey data. Their high MAE and RMSE values, coupled with lower JS Distance scores, underscore their struggles in capturing complex demographic-response relationships.
    \item \textbf{Naive Persona-based AI Survey} faces challenges in balancing variability and accuracy. With higher error metrics (MAE = 9.02, RMSE = 10.28) and moderate Cramér’s V (0.73), it highlights the trade-offs in using an unconstrained Persona-based approach.
\end{itemize}

\subsection{Comparative Visualization of Walking Preferences}
\textbf{Figure \ref{fig:survey_plots}} illustrates the walking preferences across all age groups for both the real survey and synthetic methods. Subplot (a) represents the real survey, while (b)--(g) correspond to the six synthetic surveys. In subplots (b)-(g) the real survey results are also presented with grey lines, to visually illustrate the relative performance of the various methods.

\section{Conclusions}
The \textbf{Guided Persona-based AI Survey} emerges as the most effective method, demonstrating superior alignment with real survey data across all metrics. Its ability to incorporate both demographic alignment and response constraints allows for nuanced and realistic synthetic data generation. The \textbf{Guided AI Survey} also performs well, offering a scalable solution with robust error and alignment metrics.

In contrast, methods such as the \textbf{Naive AI Survey} and \textbf{Structured AI Survey} fall short in capturing intricate demographic-response dependencies, limiting their utility for high-fidelity synthetic data generation. Persona-based approaches, while versatile and diverse, require further optimization to balance variability with accuracy.

Future research should explore hybrid models that integrate the strengths of guided Persona-based techniques while enhancing computational efficiency and scalability. Such advancements will further validate the use of artificial surveys in diverse domains. Additionally, expanding the application to more cases, including different transportation modes from the MiD dataset as well as incorporating insights from other relevant datasets, can provide a more comprehensive evaluation.

Incorporating advanced statistical measures into the evaluation framework highlights the strengths and weaknesses of different synthetic survey methods. The Guided Persona-based AI Survey consistently outperforms other methods in alignment with real survey data, demonstrating the power of LLMs to generate realistic synthetic surveys. Future work should explore additional metrics, dynamic constraints and broader applications across various datasets and domains \citep{10758652}.

\section*{Acknowledgements}
The authors acknowledge the MiD 2017 team for providing access to benchmark data and OpenAI for developing LLM tools.

\bibliography{references}

\newpage
\appendix
\section*{Appendix}
\subsection{MiD Statistics}

\begin{figure}[h!]
    \centering
    \includegraphics[width=1\linewidth]{MiD_2017_statistics.pdf}
    \caption{Normalized MiD 2017 dataset}
    \label{fig:MiD 2017 dataset}
\end{figure}

\end{document}